\documentclass[10pt, conference]{IEEEtran}
\IEEEoverridecommandlockouts
\usepackage{cite}
\usepackage{amsmath, amssymb, amsfonts}
\usepackage{algorithmic}
\usepackage{graphicx}
\usepackage{textcomp}
\usepackage{xcolor}
\usepackage{booktabs}
\usepackage{multirow}
\usepackage{array}
\usepackage{makecell}
\usepackage{longtable}
\usepackage{subcaption}
\usepackage{tikz}
\usepackage{pgfplots}
\pgfplotsset{compat=1.18}
\usepackage{placeins}
\usepackage{dblfloatfix}
\usepackage{url}
\usepackage{hyperref}
\usepackage{tabularx}

\def\BibTeX{{\rm B\kern-.05em{\sc i\kern-.025em b}\kern-.08em T\kern-.1667em\lower.7ex\hbox{E}\kern-.125emX}}

\begin{document}
    \title{Towards Gold-Standard Depth Estimation for Tree Branches in UAV
    Forestry: Benchmarking Deep Stereo Matching Methods}
    \author{\IEEEauthorblockN{Yida Lin, Bing Xue, Mengjie Zhang} \IEEEauthorblockA{\small \textit{Centre for Data Science and Artificial Intelligence} \\ \textit{Victoria University of Wellington, Wellington, New Zealand}\\ linyida\texttt{@}myvuw.ac.nz, bing.xue\texttt{@}vuw.ac.nz, mengjie.zhang\texttt{@}vuw.ac.nz}
    \and \IEEEauthorblockN{Sam Schofield, Richard Green} \IEEEauthorblockA{\small \textit{Department of Computer Science and Software Engineering} \\ \textit{University of Canterbury, Canterbury, New Zealand}\\ sam.schofield\texttt{@}canterbury.ac.nz, richard.green\texttt{@}canterbury.ac.nz}
    }
    \maketitle
    \vspace{-1.5em}

    \begin{abstract}
        Autonomous UAV forestry operations require robust depth estimation with strong
        cross-domain generalization, yet existing evaluations focus on urban and
        indoor scenarios, leaving a critical gap for vegetation-dense environments.
        We present the first systematic zero-shot evaluation of eight stereo
        methods spanning iterative refinement, foundation model, diffusion-based,
        and 3D CNN paradigms. All methods use officially released pretrained
        weights (trained on Scene Flow) and are evaluated on four standard
        benchmarks (ETH3D, KITTI 2012/2015, Middlebury) plus a novel 5,313-pair
        Canterbury Tree Branches dataset (1920$\times$1080). Results reveal scene-dependent
        patterns: foundation models excel on structured scenes (BridgeDepth:
        0.23~px on ETH3D; DEFOM: 4.65~px on Middlebury), while iterative methods
        show variable cross-benchmark performance (IGEV++: 0.36~px on ETH3D but
        6.77~px on Middlebury; IGEV: 0.33~px on ETH3D but 4.99~px on Middlebury).
        Qualitative evaluation on the Tree Branches dataset establishes DEFOM as
        the gold-standard baseline for vegetation depth estimation, with superior
        cross-domain consistency (consistently ranking 1st--2nd across benchmarks,
        average rank 1.75). DEFOM predictions will serve as pseudo-ground-truth
        for future benchmarking.
    \end{abstract}

    \vspace{-1em}
    \section{Introduction}

    Radiata pine (\textit{Pinus radiata}) is New Zealand's dominant plantation species,
    with forestry contributing NZ\$3.6 billion (1.3\%) to national GDP. Regular
    pruning is essential for high-quality clear-wood production, yet manual operations
    pose significant hazards including falls and chainsaw injuries. Autonomous
    UAV-based pruning offers a safer alternative~\cite{lin2024branch,steininger2025timbervision},
    but requires centimeter-level depth accuracy for precise tool positioning at
    1--2~m operating distances. Recent advances have demonstrated the potential of
    integrating object detection with stereo vision~\cite{lin2025yolosgbm}, deep
    learning for branch segmentation~\cite{lin2025segmentation}, and evolutionary
    optimization for stereo matching~\cite{lin2025genetic}. Unlike urban or
    indoor scenes, forest canopies feature thin overlapping branches, repetitive
    patterns, extreme depth discontinuities, and dramatic illumination
    variations. Stereo vision provides a passive, lightweight alternative to
    power-hungry active sensors, ideal for resource-constrained aerial platforms.

    Deep learning has advanced stereo matching significantly through end-to-end trainable
    architectures. The Scene Flow dataset~\cite{mayer2016large} has become the
    de facto standard for pretraining, providing over 39,000 synthetic stereo pairs
    with dense, pixel-perfect ground-truth disparity---annotations that are
    impractical to obtain at scale in real-world scenes. Our goal is to identify
    the most robust pretrained model for generating pseudo-ground-truth on our Tree
    Branches dataset, which comprises 5,313 high-quality stereo pairs captured
    using a ZED Mini camera mounted on a UAV. Since no LiDAR annotation exists,
    we evaluate models through zero-shot inference on established benchmarks---KITTI~\cite{geiger2012kitti},
    Middlebury~\cite{scharstein2014middlebury}, and ETH3D~\cite{schops2017eth3d}---using
    Scene Flow pretrained weights without any fine-tuning. Based on cross-domain
    generalization performance, we then select the most robust method to
    generate gold-standard pseudo-ground-truth for our Tree Branches dataset. Inference
    speed is not considered, as our focus is purely on prediction quality.

    Three gaps motivate this study. \textit{First}, existing stereo benchmarks (KITTI,
    ETH3D, Middlebury) focus on automotive and indoor scenarios, providing
    limited insight for vegetation-dense environments. \textit{Second}, most comparative
    studies permit domain-specific fine-tuning, obscuring the pure
    generalization capability essential when target-domain annotations---like
    LiDAR depth maps in dense canopies---are unavailable. \textit{Third}, the relative
    performance of recent paradigms---iterative refinement, foundation models, and
    diffusion-based methods---remains unclear for cross-domain deployment in forestry
    applications.

    To address these gaps, we evaluate eight deep stereo methods spanning four
    architectural paradigms: iterative refinement (RAFT-Stereo~\cite{lipson2021raft},
    IGEV~\cite{xu2023igev}, IGEV++~\cite{xu2024igevpp}), foundation models (BridgeDepth~\cite{guan2025bridgedepth},
    DEFOM-Stereo~\cite{jiang2025defom}, hereafter DEFOM), diffusion-based (StereoAnywhere~\cite{zhao2024stereoanywhere}),
    and 3D CNN (ACVNet~\cite{xu2022acvnet}, PSMNet~\cite{chang2018psmnet}). All
    methods use Scene Flow pretrained weights and are evaluated zero-shot on
    four standard benchmarks plus our 5,313-pair Tree Branches dataset. Our contributions
    are:

    \begin{itemize}
        \item \textbf{Systematic paradigm comparison}: The first zero-shot evaluation
            comparing iterative refinement, foundation models, diffusion-based,
            and 3D CNN methods under identical training conditions to isolate
            inherent generalization capability for forestry deployment.

        \item \textbf{Comprehensive cross-domain evaluation}: Consistent assessment
            on four standard benchmarks spanning indoor, automotive, and high-resolution
            scenarios.

        \item \textbf{Performance analysis}: Quantitative comparison revealing scene-dependent
            patterns---foundation models excel on structured scenes while iterative
            methods show variable cross-benchmark robustness---informing method selection
            for vegetation-specific applications.

        \item \textbf{Gold-standard baseline for forestry}: DEFOM identified as optimal
            for the Tree Branches dataset based on superior cross-domain consistency
            (average rank 1.75 across benchmarks), enabling future quantitative
            benchmarking with pseudo-ground-truth depth maps.
    \end{itemize}
    \vspace{-1em}
    \section{Related Work}

    \subsection{Deep Stereo Matching Architectures}

    Deep learning has transformed stereo matching through end-to-end trainable architectures.
    Early 3D CNN methods established the foundation: DispNet~\cite{mayer2016large}
    demonstrated end-to-end disparity regression, GC-Net~\cite{kendall2017gcnet}
    introduced 3D convolutions for cost volume regularization, and PSMNet~\cite{chang2018psmnet}
    incorporated spatial pyramid pooling for multi-scale context aggregation.
    ACVNet~\cite{xu2022acvnet} advanced this paradigm with attention-based cost
    volume construction. While effective on training domains, these methods
    struggle with large disparity ranges and cross-domain transfer---a critical
    limitation for forestry applications where Scene Flow training data differs
    substantially from vegetation scenes.

    Iterative refinement methods, inspired by optical flow estimation, address
    these limitations through progressive disparity updates. RAFT-Stereo~\cite{lipson2021raft}
    adapts recurrent all-pairs field transforms with multi-scale correlation volumes.
    IGEV~\cite{xu2023igev} combines iterative updates with geometry encoding
    volumes to handle occlusions and textureless regions. IGEV++~\cite{xu2024igevpp}
    extends this with multi-range geometry encoding for large disparities while preserving
    fine-grained details---potentially beneficial for capturing thin branch structures
    in forestry scenes.

    \subsection{Foundation Models and Diffusion-Based Methods}

    Recent foundation model approaches exploit large-scale pre-training and
    monocular depth priors to enhance cross-domain generalization. BridgeDepth~\cite{guan2025bridgedepth}
    bridges monocular contextual reasoning and stereo geometric matching through
    iterative bidirectional alignment in latent space. DEFOM-Stereo~\cite{jiang2025defom}
    incorporates Depth Anything V2 into recurrent stereo matching with scale
    update modules, leveraging monocular depth priors trained on diverse real-world
    scenes. These methods are particularly promising for forestry applications
    where monocular cues can resolve ambiguities in textureless foliage regions.

    Diffusion-based paradigms offer an alternative approach to cross-domain
    transfer. StereoAnywhere~\cite{zhao2024stereoanywhere} formulates stereo matching
    as conditional image generation, enabling zero-shot transfer through learned
    image priors. However, systematic comparison of these paradigms---iterative
    refinement, foundation models, and diffusion-based methods---under identical
    zero-shot conditions remains limited, particularly for vegetation-dense environments.

    \subsection{Stereo Vision for UAV Forestry}

    UAV-based stereo vision has been explored for navigation~\cite{fraundorfer2012vision},
    3D reconstruction~\cite{nex2014uav}, and obstacle avoidance~\cite{barry2015pushbroom}.
    However, most systems target structured urban environments, leaving forestry
    applications underexplored. Forest canopies present unique challenges for
    stereo matching: thin overlapping branches require fine-grained depth resolution,
    repetitive foliage patterns create ambiguous correspondences, extreme depth discontinuities
    challenge cost aggregation, and variable natural illumination causes
    appearance changes between stereo views.

    While monocular depth has been applied to forest inventory~\cite{jayathunga2018forest},
    stereo-based approaches offer geometric constraints essential for precise depth
    estimation at close range (1--2~m) required for autonomous pruning
    operations. Recent work on branch detection~\cite{lin2024branch,lin2025yolosgbm}
    and segmentation~\cite{lin2025segmentation} has demonstrated the potential of
    integrating deep learning with stereo vision for forestry, but relies on
    classical matching methods. Evaluating modern deep stereo architectures for vegetation
    scenes remains an open challenge.

    \subsection{Zero-Shot Cross-Domain Generalization}

    Domain adaptation in stereo matching typically employs supervised fine-tuning~\cite{tonioni2019domain}
    or self-supervised adaptation~\cite{watson2020self}, but requires target-domain
    data---often impractical for forestry where dense ground-truth acquisition via
    LiDAR is hindered by canopy occlusion. Zero-shot generalization---deploying
    models trained solely on synthetic data---offers a practical alternative.

    The Scene Flow dataset~\cite{mayer2016large} has become the de facto
    standard for pre-training, providing dense ground-truth unavailable at scale
    in real-world scenes. Existing benchmarks (KITTI~\cite{geiger2012kitti},
    ETH3D~\cite{schops2017eth3d}, Middlebury~\cite{scharstein2014middlebury})
    enable cross-domain evaluation but focus on automotive and indoor scenarios,
    providing limited insight for vegetation-dense environments. Our work
    addresses this gap through systematic zero-shot evaluation across diverse
    benchmarks plus a novel forestry dataset, identifying optimal methods for generating
    pseudo-ground-truth where LiDAR annotation is infeasible.

    \section{Methodology}

    \subsection{Problem Formulation}

    Stereo matching estimates per-pixel disparity from a rectified stereo image pair.
    Given left and right images
    $I_{L}, I_{R}\in \mathbb{R}^{H \times W \times 3}$, the goal is to compute a
    disparity map $D \in \mathbb{R}^{H \times W}$ where each pixel $(x,y)$ in the
    left image corresponds to pixel $(x - D(x,y), y)$ in the right image. The
    depth $Z$ at pixel $(x,y)$ can be recovered through triangulation:
    \begin{equation}
        Z(x,y) = \frac{f \cdot B}{D(x,y)}
    \end{equation}
    where $f$ is the focal length and $B$ is the stereo baseline. Deep stereo
    methods learn a mapping
    $f_{\theta}: \mathbb{R}^{H \times W \times 3}\times \mathbb{R}^{H \times W
    \times 3}\rightarrow \mathbb{R}^{H \times W}$
    parameterized by weights $\theta$, trained to minimize disparity prediction error
    on labeled data.

    \subsection{Datasets}

    We evaluate stereo matching methods across five diverse datasets to assess
    generalization capability:

    \textbf{Scene Flow}~\cite{mayer2016large} provides over 39,000 synthetic stereo
    pairs with perfect ground-truth for training.

    \textbf{ETH3D}~\cite{schops2017eth3d} contains high-resolution indoor/outdoor
    pairs with structured-light ground-truth.

    \textbf{KITTI 2012/2015}~\cite{geiger2012kitti,menze2015kitti} represent autonomous
    driving scenarios with LiDAR ground-truth.

    \textbf{Middlebury}~\cite{scharstein2014middlebury} provides high-accuracy
    indoor ground-truth for fine-grained disparity estimation.

    \textbf{Tree Branches Dataset}: We introduce 5,313 stereo pairs (1920$\times$1080)
    captured using a ZED Mini camera (63~mm baseline) mounted on a UAV in
    Canterbury, New Zealand (March--October 2024). Pairs were rigorously selected
    from hundreds of thousands of raw captures through multi-stage filtering assessing
    motion blur, exposure quality, stereo rectification accuracy, and scene
    diversity. The dataset targets vegetation-specific challenges: thin
    overlapping branches, repetitive foliage patterns, extreme depth
    discontinuities, and variable illumination. \textit{No ground-truth depth
    maps are provided}. The dataset will be publicly released with DEFOM pseudo-ground-truth
    to facilitate UAV forestry research.

    \subsection{Evaluated Methods}

\begin{figure*}[htbp]
        \centering
        \includegraphics[width=0.8\linewidth]{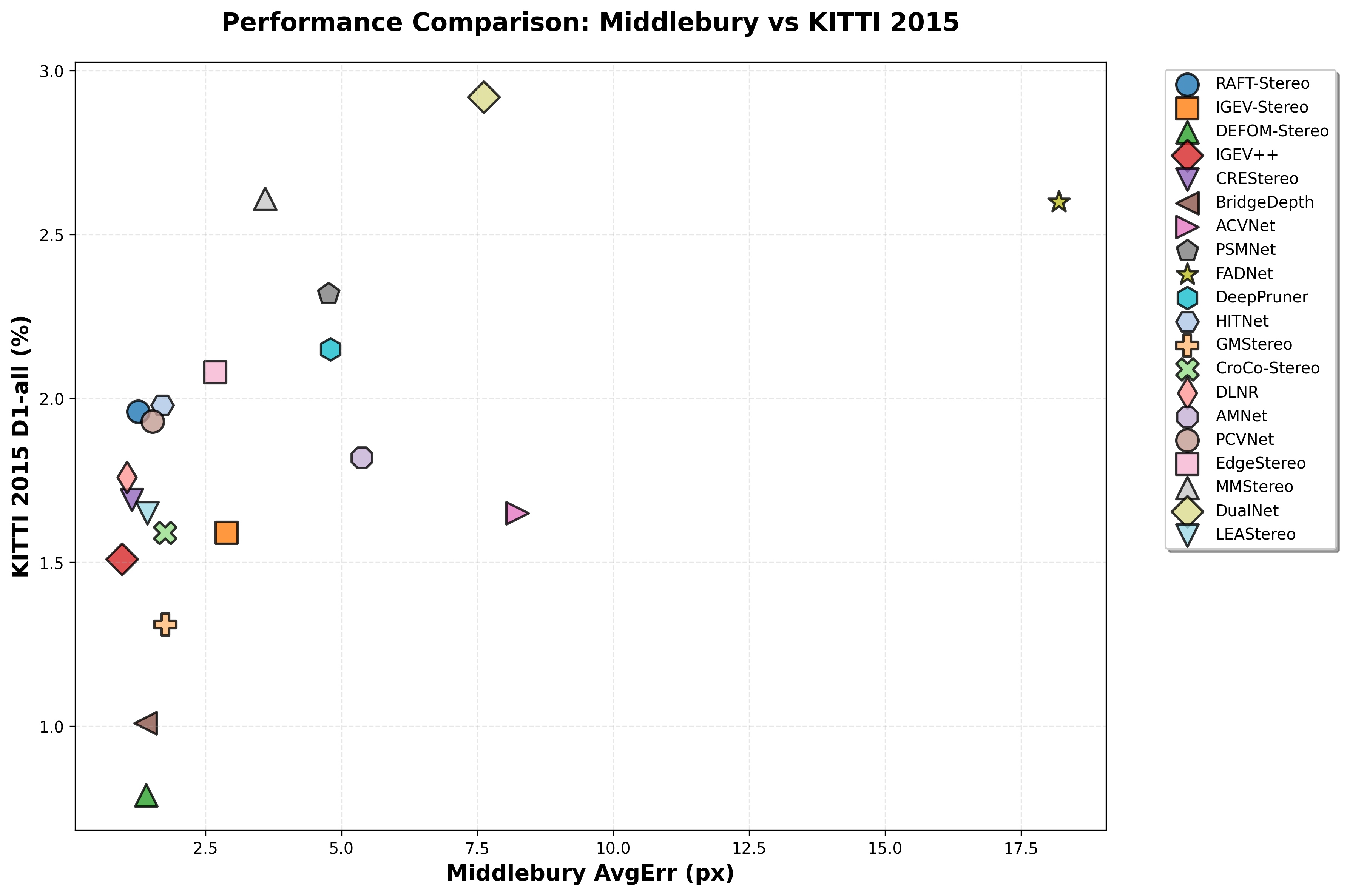}
        \caption{Initial screening of 20 stereo matching methods using
        officially released pretrained weights on KITTI 2015 (D1-all \%) and
        Middlebury (Average Absolute Error, pixels). Foundation models (DEFOM:
        0.79\% D1, BridgeDepth: 1.01\% D1) dominate KITTI 2015, while iterative
        methods (IGEV++: 0.97~px AAE, DLNR: 1.06~px) excel on Middlebury. Based
        on this screening, eight methods (highlighted) are selected for comprehensive
        cross-domain generalization evaluation, balancing performance and
        architectural diversity. Classical 3D CNN methods (ACVNet, PSMNet) are
        included as baselines despite higher errors.}
        \label{fig:performance_comparison}
    \end{figure*}

    We first survey 20 representative deep stereo methods using their officially
    released pretrained weights and evaluate them on KITTI 2015 and Middlebury benchmarks
    (Fig.~\ref{fig:performance_comparison}). Based on this initial screening, we
    select eight methods spanning different architectural paradigms for
    comprehensive cross-domain generalization evaluation (Table~\ref{tab:methods}).
    Selection criteria include: (1) competitive performance on at least one
    benchmark, (2) coverage of diverse architectural paradigms, and (3)
    availability of Scene Flow pretrained weights for fair comparison.

    \begin{table}[htbp]
        \caption{Evaluated Stereo Matching Methods and Their Architectural
        Paradigms}
        \label{tab:methods}
        \centering
        \small
        \begin{tabular}{lcc}
            \toprule \textbf{Method}                     & \textbf{Type} & \textbf{Key Feature}       \\
            \midrule RAFT-Stereo~\cite{lipson2021raft}   & Iterative     & Recurrent field transforms \\
            IGEV~\cite{xu2023igev}                       & Iterative     & Geometry encoding volume   \\
            IGEV++~\cite{xu2024igevpp}                   & Iterative     & Multi-range encoding       \\
            BridgeDepth~\cite{guan2025bridgedepth}       & Foundation    & Latent alignment           \\
            DEFOM~\cite{jiang2025defom}                  & Foundation    & Depth foundation model     \\
            StereoAnywhere~\cite{zhao2024stereoanywhere} & Diffusion     & Zero-shot transfer         \\
            \midrule ACVNet~\cite{xu2022acvnet}          & 3D CNN        & Attention concatenation    \\
            PSMNet~\cite{chang2018psmnet}                & 3D CNN        & Spatial pyramid pooling    \\
            \bottomrule
        \end{tabular}
    \end{table}

    \subsection{Evaluation Metrics}

    We employ two complementary metrics that capture different aspects of stereo
    matching quality. \textbf{End-Point Error (EPE)} measures the average
    absolute disparity error:
    \begin{equation}
        \text{EPE}= \frac{1}{N}\sum_{i=1}^{N}|d_{i}- \hat{d}_{i}|
    \end{equation}
    EPE reflects overall prediction accuracy and is sensitive to systematic bias,
    making it suitable for assessing methods intended as pseudo-ground-truth
    baselines. \textbf{D1-Error} computes the percentage of pixels with error
    exceeding 3 pixels:
    \begin{equation}
        \text{D1}= \frac{1}{N}\sum_{i=1}^{N}\mathbb{1}(|d_{i}- \hat{d}_{i}| > 3)
    \end{equation}
    D1 captures the proportion of catastrophic failures independent of error magnitude,
    critical for safety applications where consistent reliability matters more
    than mean accuracy.

    For UAV forestry, EPE directly relates to depth accuracy: sub-pixel
    disparity errors translate to centimeter-level depth precision. D1 identifies
    regions of complete matching failure common in vegetation scenes. Together, these
    metrics distinguish methods with consistent performance from those with high
    failure rates.

    \subsection{Implementation Details}

    \textbf{Pre-trained Models}: All methods use officially released Scene Flow
    weights~\cite{mayer2016large} with zero-shot inference.

    \textbf{Evaluation Protocol}: Experiments are conducted on an NVIDIA Quadro
    RTX 6000 GPU (24~GB VRAM) with PyTorch 2.6.0. We use full-resolution images
    without cropping. Invalid pixels are excluded following standard protocols.
    For stochastic methods, we report means across three runs.

    \section{Experimental Results}

    \subsection{Performance on Standard Benchmarks}

    Table~\ref{tab:benchmark_results} summarizes zero-shot results on four standard
    benchmarks.

    \begin{table*}
        [htbp]
        \caption{Cross-Domain Generalization Performance on Standard Benchmarks}
        \label{tab:benchmark_results}
        \centering
        \small
        \begin{tabular}{lcccccccc}
            \toprule \multirow{2}{*}{\textbf{Method}}                                   & \multicolumn{2}{c}{\textbf{ETH3D}} & \multicolumn{2}{c}{\textbf{KITTI 2012}} & \multicolumn{2}{c}{\textbf{KITTI 2015}} & \multicolumn{2}{c}{\textbf{Middlebury}} \\
            \cmidrule(lr){2-3} \cmidrule(lr){4-5} \cmidrule(lr){6-7} \cmidrule(lr){8-9} & EPE$\downarrow$                    & D1$\downarrow$                          & EPE$\downarrow$                         & D1$\downarrow$                         & EPE$\downarrow$ & D1$\downarrow$ & EPE$\downarrow$ & D1$\downarrow$ \\
            \midrule RAFT-Stereo                                                        & 0.27                               & 0.88                                    & 0.90                                    & 4.41                                   & 1.11            & 5.12           & 5.50            & 10.80          \\
            IGEV                                                                        & 0.33                               & 1.44                                    & 1.03                                    & 5.21                                   & 1.17            & 5.45           & 4.99            & \textbf{6.79}  \\
            IGEV++                                                                      & 0.36                               & 1.70                                    & 1.20                                    & 6.37                                   & 1.23            & 5.83           & 6.77            & 7.82           \\
            BridgeDepth                                                                 & \textbf{0.23}                      & \textbf{0.39}                           & \textbf{0.83}                           & \textbf{3.65}                          & 1.07            & \textbf{4.34}  & 20.03           & 19.54          \\
            StereoAnywhere                                                              & 0.43                               & 2.04                                    & 1.02                                    & 4.91                                   & 1.11            & 5.43           & 9.51            & 18.84          \\
            DEFOM                                                                       & 0.35                               & 0.92                                    & 0.84                                    & 3.76                                   & \textbf{1.04}   & 4.57           & \textbf{4.65}   & 8.28           \\
            \midrule ACVNet                                                             & 1.95                               & 3.50                                    & 1.91                                    & 11.72                                  & 2.18            & 9.95           & 37.36           & 36.67          \\
            PSMNet                                                                      & 2.15                               & 4.20                                    & 3.77                                    & 27.32                                  & 3.97            & 28.21          & 48.62           & 54.42          \\
            \bottomrule
        \end{tabular}
        \vspace{0.5em}
        \begin{flushleft}
            \footnotesize{EPE: End-Point Error (pixels), D1: Percentage of pixels with error $>$3~px (\%). All methods use Scene Flow pretrained weights without fine-tuning. Bold indicates best per benchmark.}
        \end{flushleft}
    \end{table*}

    \subsubsection{Performance Analysis by Method}

    \textbf{Foundation Models (BridgeDepth, DEFOM)}: Both methods leverage monocular
    depth priors from large-scale pretraining. BridgeDepth achieves best results
    on ETH3D (0.23~px EPE, 0.39\% D1) and KITTI (0.83~px on KITTI 2012, 3.65\% D1),
    but degrades significantly on Middlebury (20.03~px EPE, 19.54\% D1). This
    suggests BridgeDepth's latent alignment works well within moderate disparity
    ranges but struggles with Middlebury's extreme disparities. DEFOM shows more
    balanced performance: competitive on ETH3D (0.35~px) and KITTI (0.84~px on KITTI
    2012, 1.04~px on KITTI 2015), while achieving best EPE on Middlebury (4.65~px).
    DEFOM's scale update modules enable adaptation to varying disparity ranges.

    \textbf{Iterative Methods (RAFT-Stereo, IGEV, IGEV++)}: These methods show moderate
    and consistent performance. RAFT-Stereo achieves 0.27~px on ETH3D and 0.90~px
    on KITTI 2012. IGEV performs similarly (0.33~px on ETH3D, 1.03~px on KITTI
    2012) but achieves best D1 on Middlebury (6.79\%). IGEV++ shows slightly higher
    errors (0.36~px on ETH3D, 1.20~px on KITTI 2012), indicating that additional
    complexity does not always improve generalization.

    \textbf{Diffusion-Based (StereoAnywhere)}: Moderate performance across benchmarks
    (0.43~px on ETH3D, 1.02~px on KITTI 2012, 9.51~px on Middlebury). The
    diffusion formulation provides reasonable zero-shot transfer but does not
    outperform foundation models or iterative methods.

    \textbf{Classical 3D CNN (ACVNet, PSMNet)}: Both methods fail significantly
    across all benchmarks. ACVNet shows 1.95~px on ETH3D and 37.36~px on Middlebury.
    PSMNet performs worst: 2.15~px on ETH3D, 3.77~px on KITTI 2012, and 48.62~px
    on Middlebury with 54.42\% D1. These failures stem from fixed disparity range
    assumptions (e.g., PSMNet's 192-pixel limit) and lack of semantic priors,
    making them unsuitable for cross-domain deployment.

    \subsubsection{Method Selection: DEFOM as the Most Robust Choice}

    Among all methods, \textbf{DEFOM demonstrates the most stable cross-domain
    performance}. Comparing EPE rankings across benchmarks: DEFOM ranks 4th on ETH3D
    (0.35~px), 2nd on KITTI 2012 (0.84~px), 1st on KITTI 2015 (1.04~px), and 1st
    on Middlebury (4.65~px). While BridgeDepth achieves lower errors on ETH3D and
    KITTI, its Middlebury failure (20.03~px vs.\ DEFOM's 4.65~px) reveals poor generalization
    to large disparity ranges. IGEV achieves best Middlebury D1 (6.79\%) but shows
    higher EPE on KITTI datasets. DEFOM's consistent top-tier performance across
    diverse scenarios---without catastrophic failures on any benchmark---makes
    it the safest choice for generating pseudo-ground-truth on unseen domains
    like our Tree Branches dataset.

    \begin{figure*}[htbp]
        \centering
        \begin{subfigure}
            [b]{0.24\textwidth}
            \includegraphics[width=\textwidth]{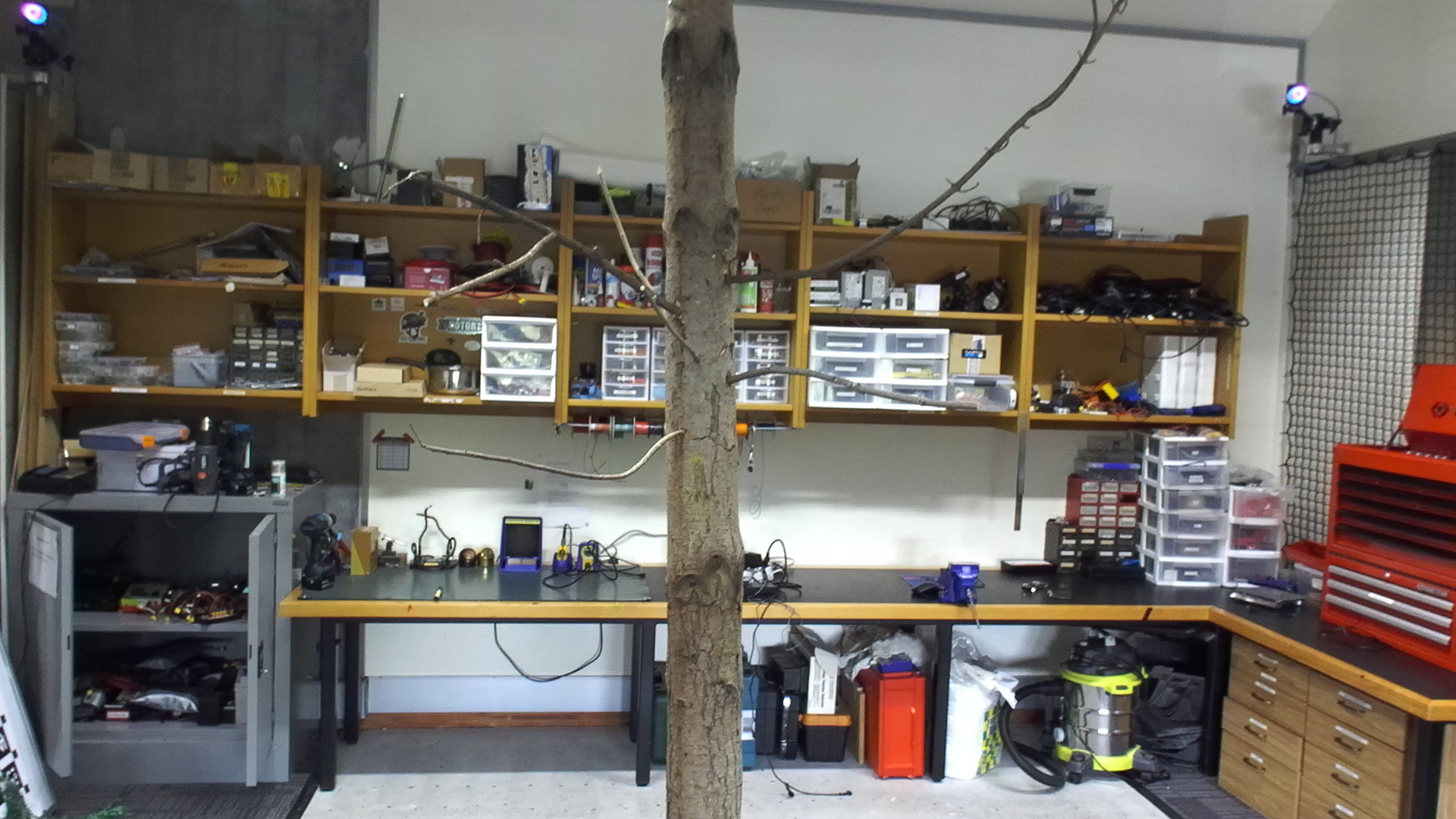}
            \caption{Left image}
        \end{subfigure}
        \hfill
        \begin{subfigure}
            [b]{0.24\textwidth}
            \includegraphics[width=\textwidth]{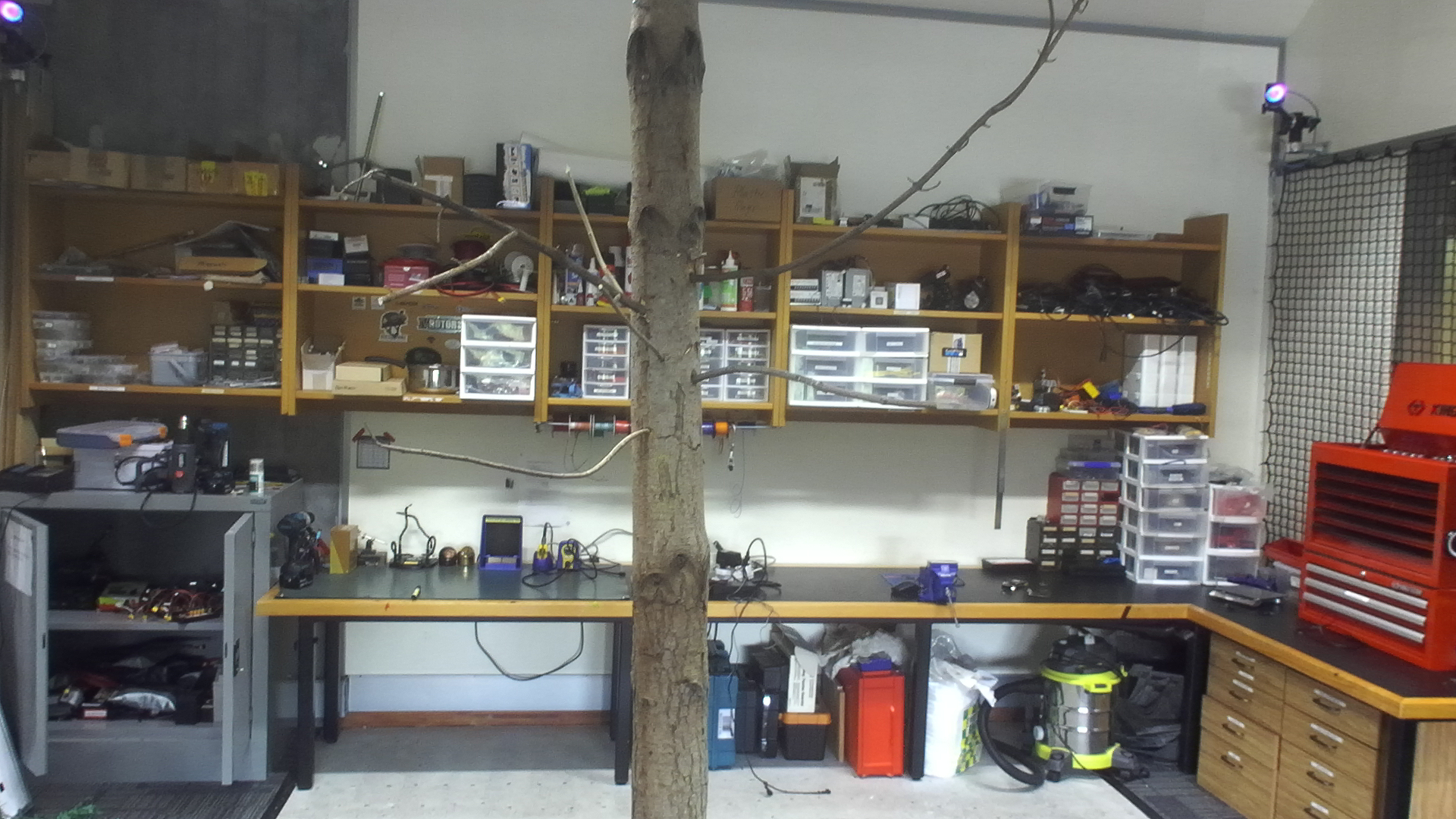}
            \caption{Right image}
        \end{subfigure}
        \hfill
        \begin{subfigure}
            [b]{0.24\textwidth}
            \includegraphics[width=\textwidth]{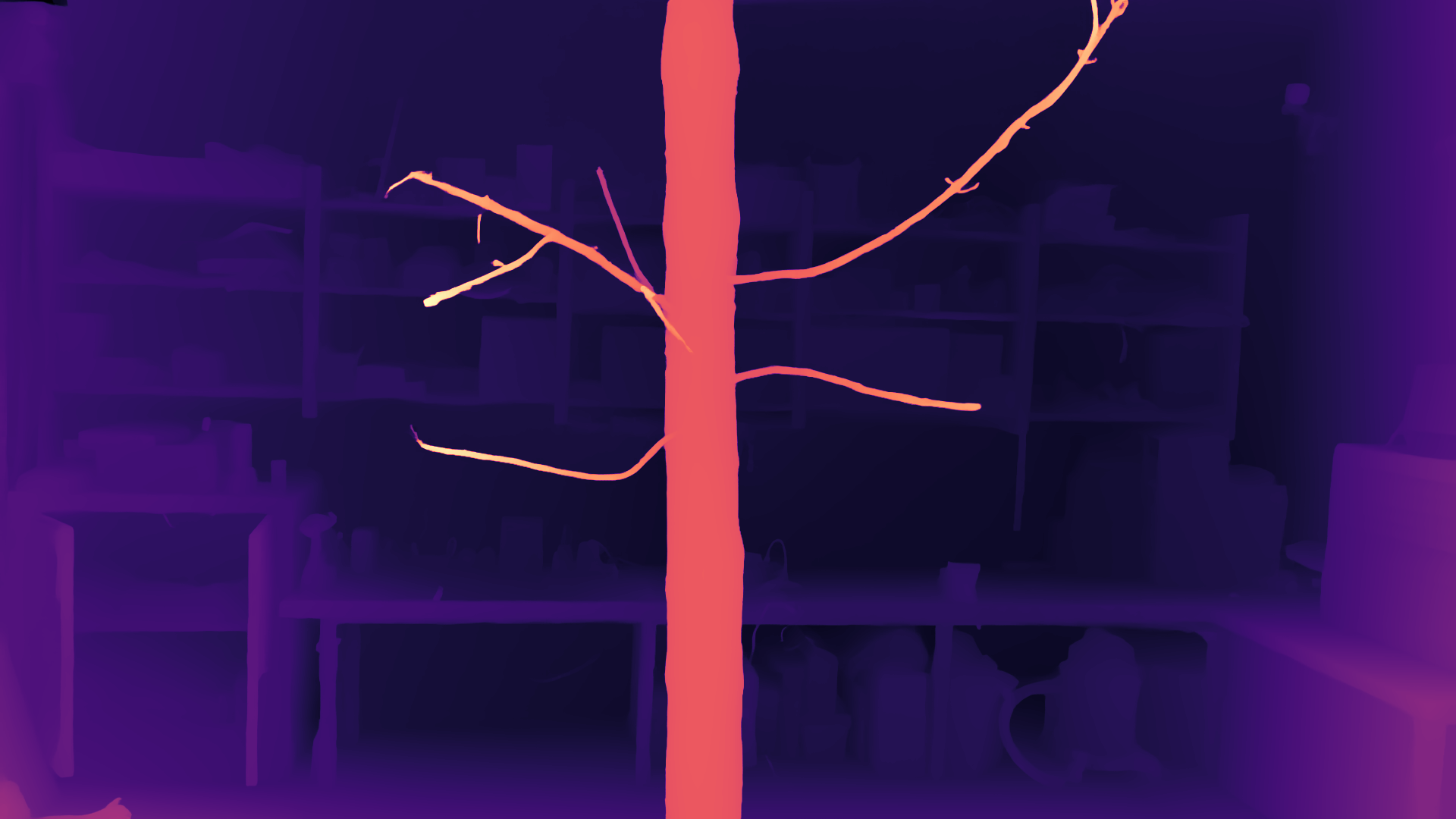}
            \caption{DEFOM prediction}
        \end{subfigure}
        \hfill
        \begin{subfigure}
            [b]{0.24\textwidth}
            \includegraphics[width=\textwidth]{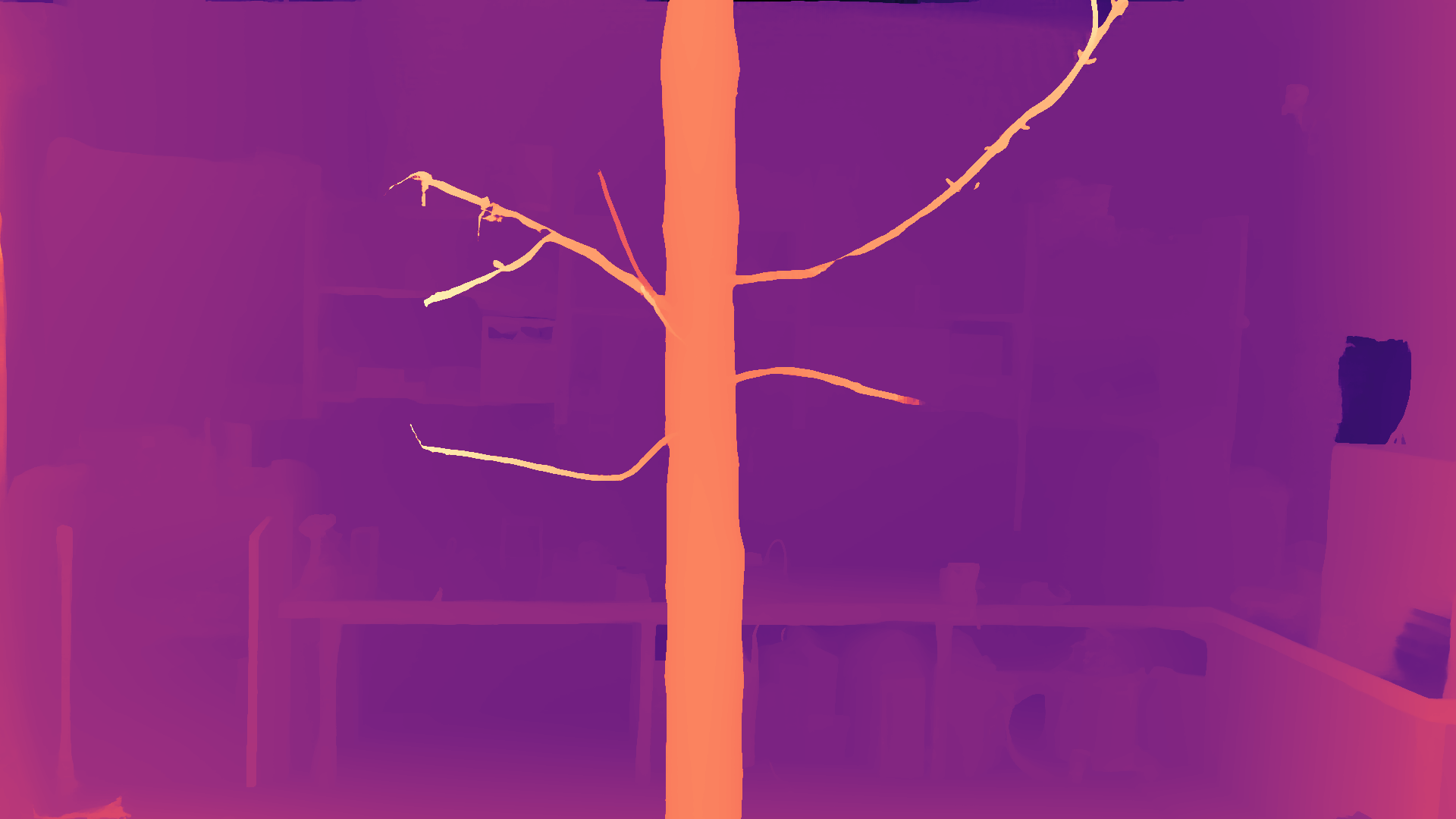}
            \caption{IGEV++ prediction}
        \end{subfigure}
        \caption{Qualitative comparison on Scene 3305. DEFOM produces smoother
        disparity maps with better sky-region consistency. IGEV++ preserves finer
        branch details but exhibits noise in homogeneous areas.}
        \label{fig:scene_3305}
    \end{figure*}

    \begin{figure*}[htbp]
        \centering
        \begin{subfigure}
            [b]{0.24\textwidth}
            \includegraphics[width=\textwidth]{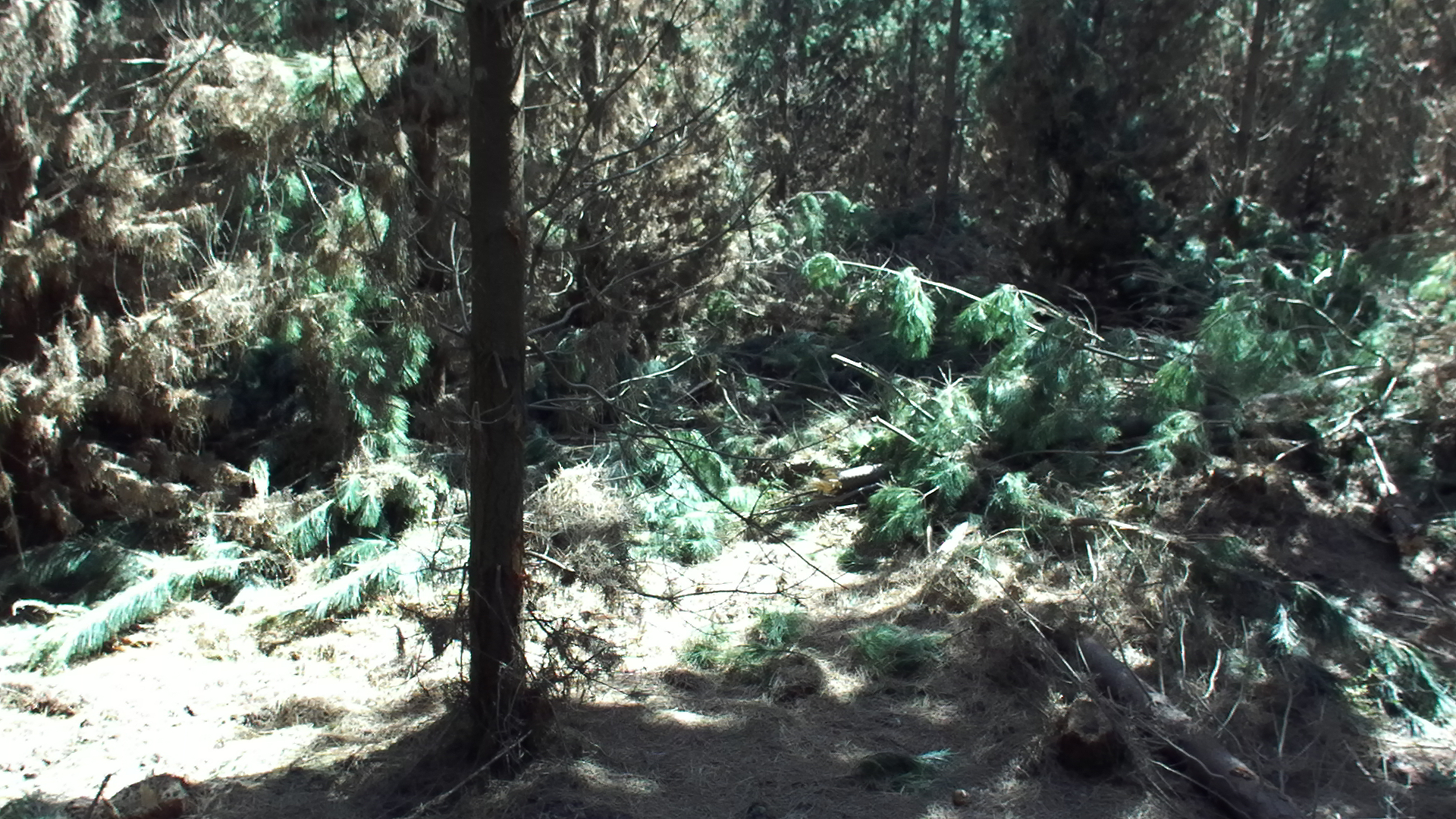}
            \caption{Left image}
        \end{subfigure}
        \hfill
        \begin{subfigure}
            [b]{0.24\textwidth}
            \includegraphics[width=\textwidth]{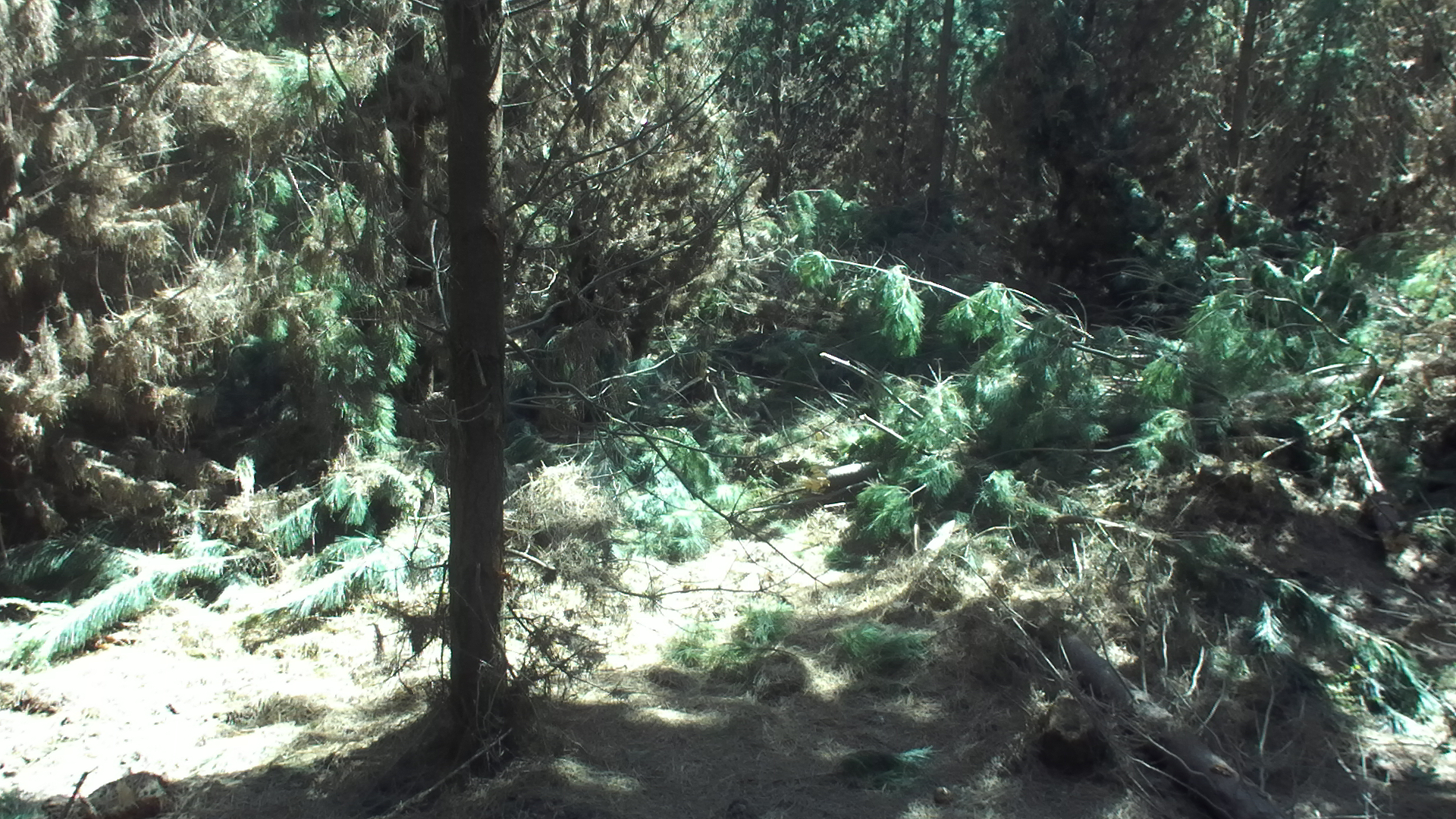}
            \caption{Right image}
        \end{subfigure}
        \hfill
        \begin{subfigure}
            [b]{0.24\textwidth}
            \includegraphics[width=\textwidth]{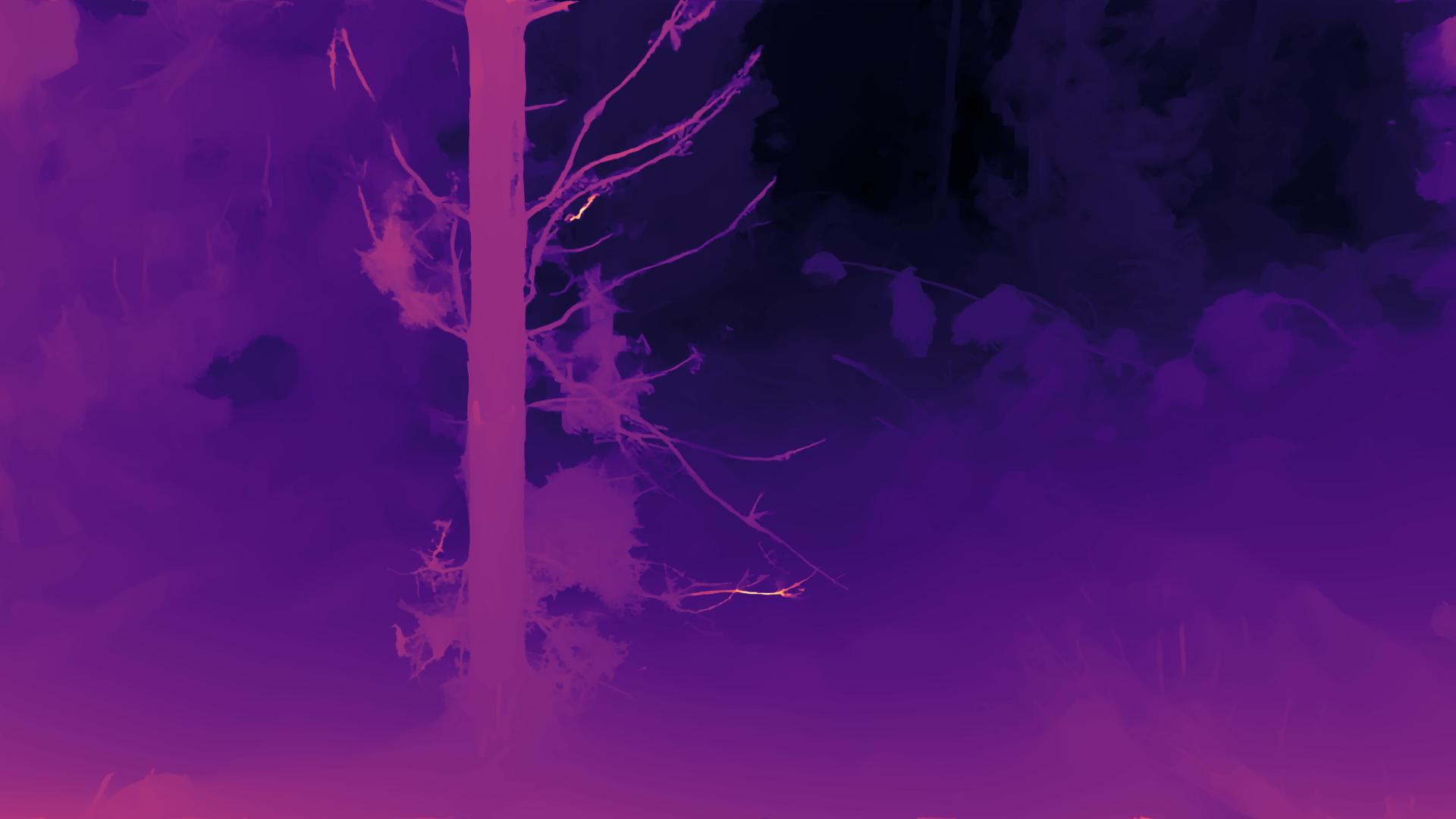}
            \caption{DEFOM prediction}
        \end{subfigure}
        \hfill
        \begin{subfigure}
            [b]{0.24\textwidth}
            \includegraphics[width=\textwidth]{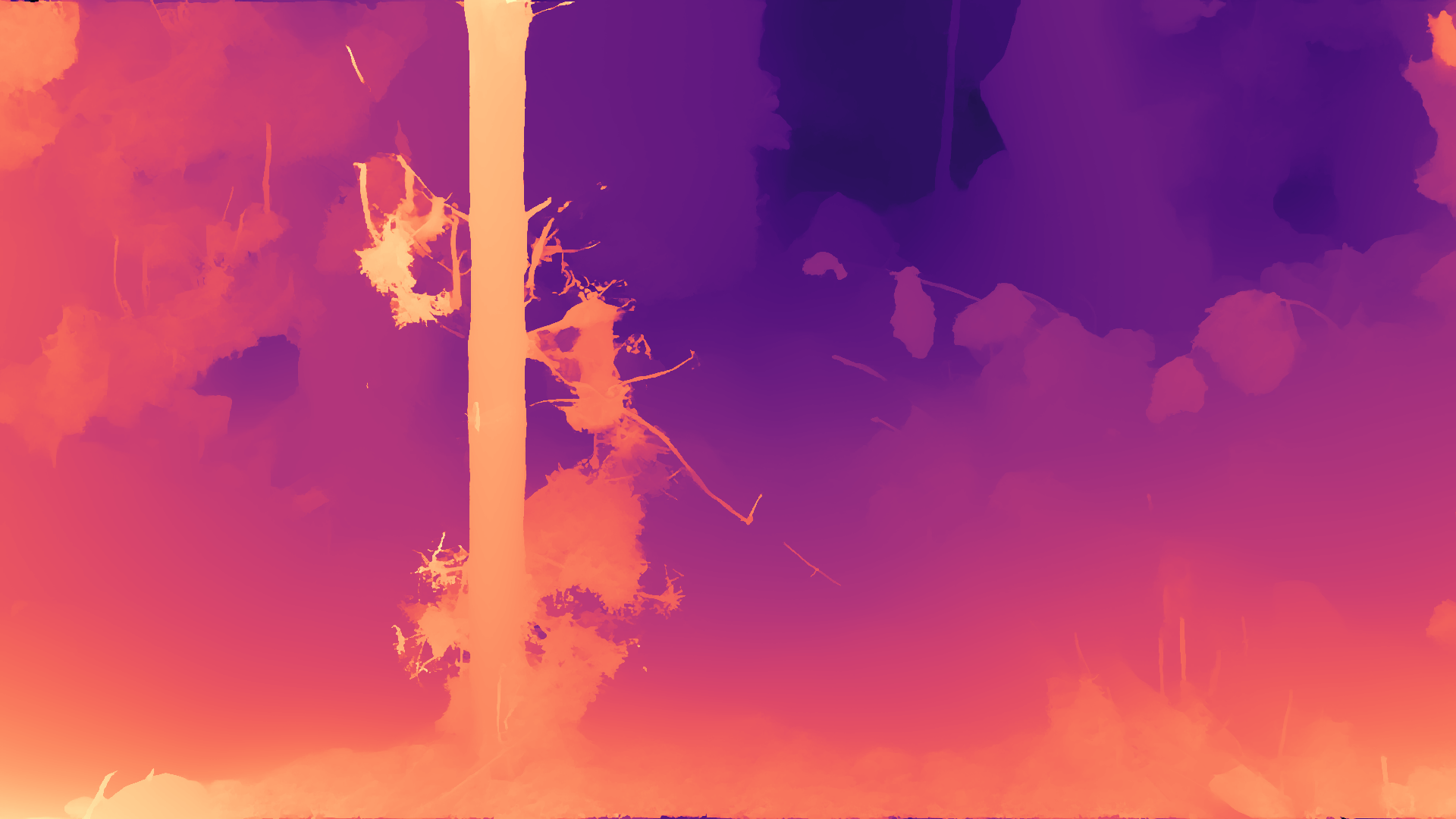}
            \caption{IGEV++ prediction}
        \end{subfigure}
        \caption{Qualitative comparison on Scene 4939. DEFOM generates smooth
        depth transitions at occlusion boundaries. IGEV++ produces sharper edges
        but exhibits artifacts near thin structures.}
        \label{fig:scene_4939}
    \end{figure*}

    \begin{figure*}[htbp]
        \centering
        \begin{subfigure}
            [b]{0.24\textwidth}
            \includegraphics[width=\textwidth]{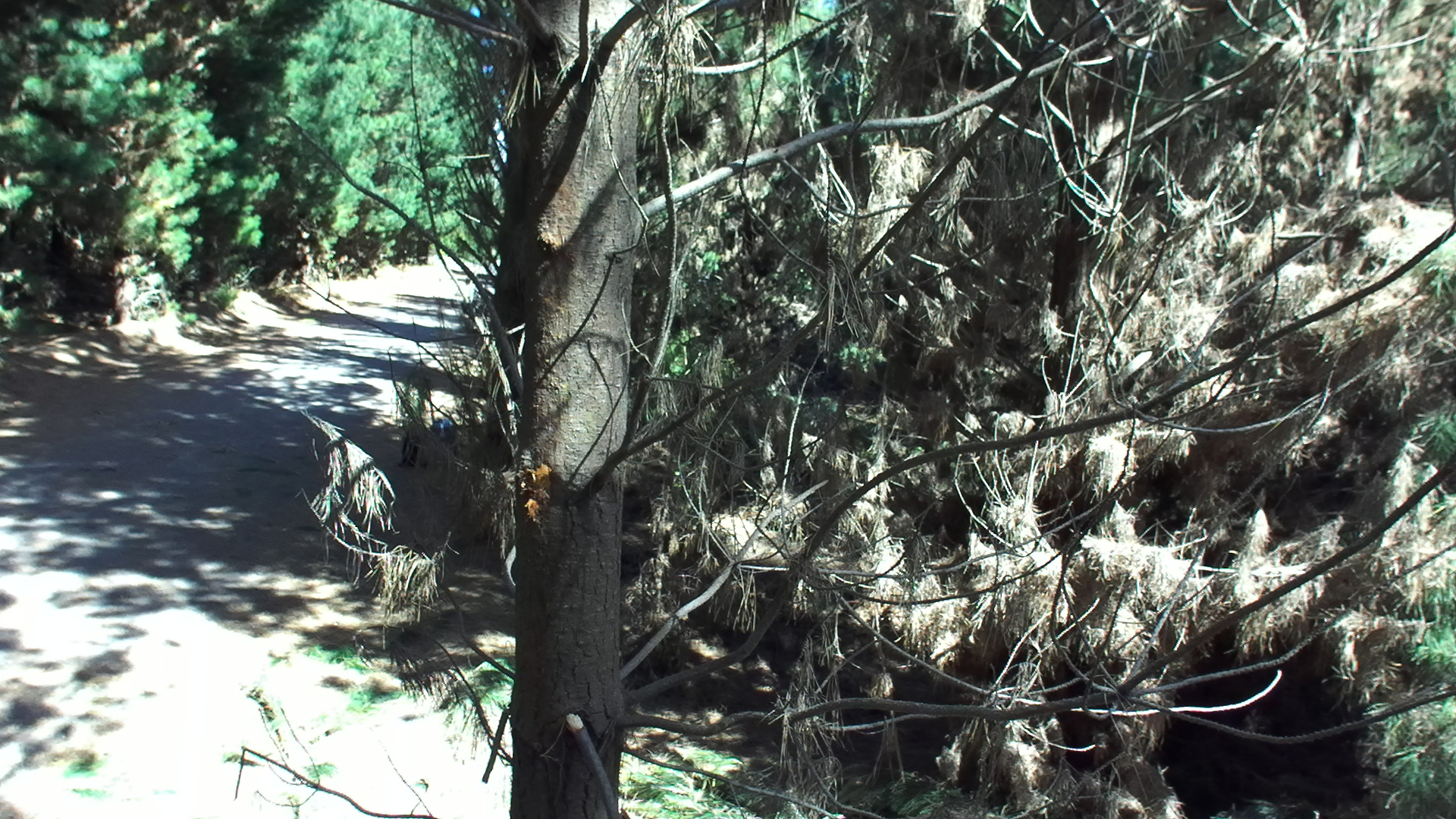}
            \caption{Left image}
        \end{subfigure}
        \hfill
        \begin{subfigure}
            [b]{0.24\textwidth}
            \includegraphics[width=\textwidth]{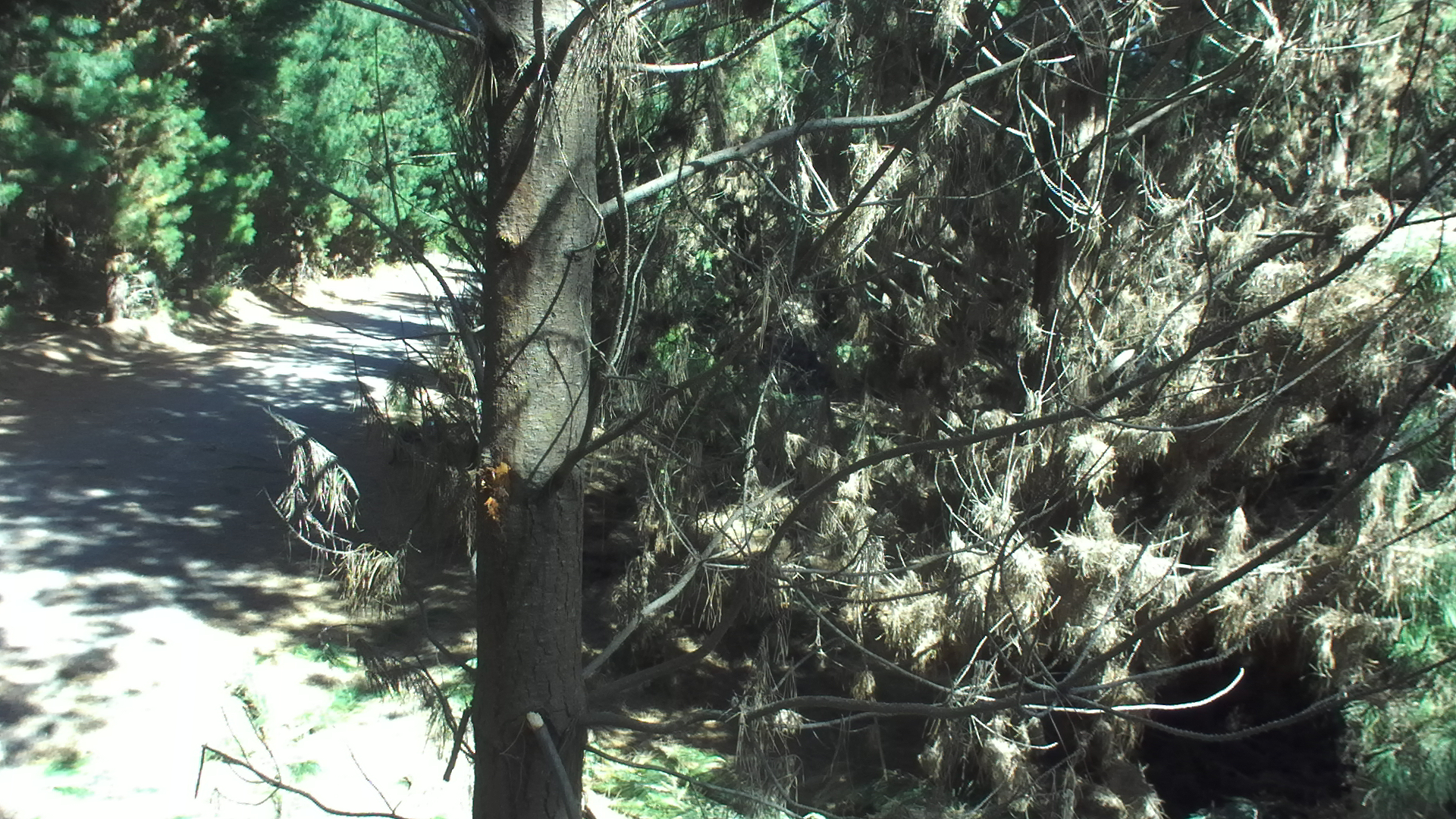}
            \caption{Right image}
        \end{subfigure}
        \hfill
        \begin{subfigure}
            [b]{0.24\textwidth}
            \includegraphics[width=\textwidth]{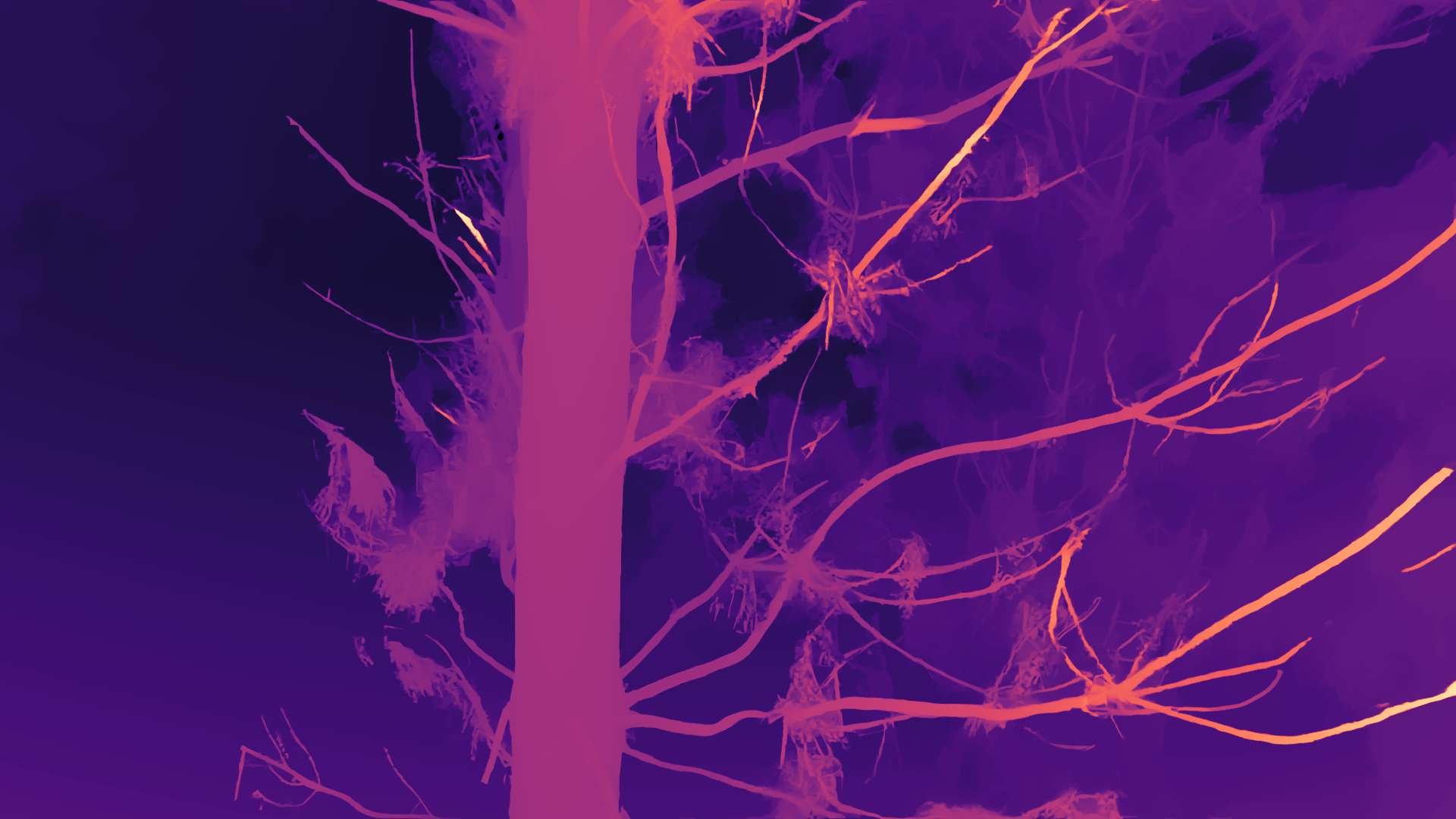}
            \caption{DEFOM prediction}
        \end{subfigure}
        \hfill
        \begin{subfigure}
            [b]{0.24\textwidth}
            \includegraphics[width=\textwidth]{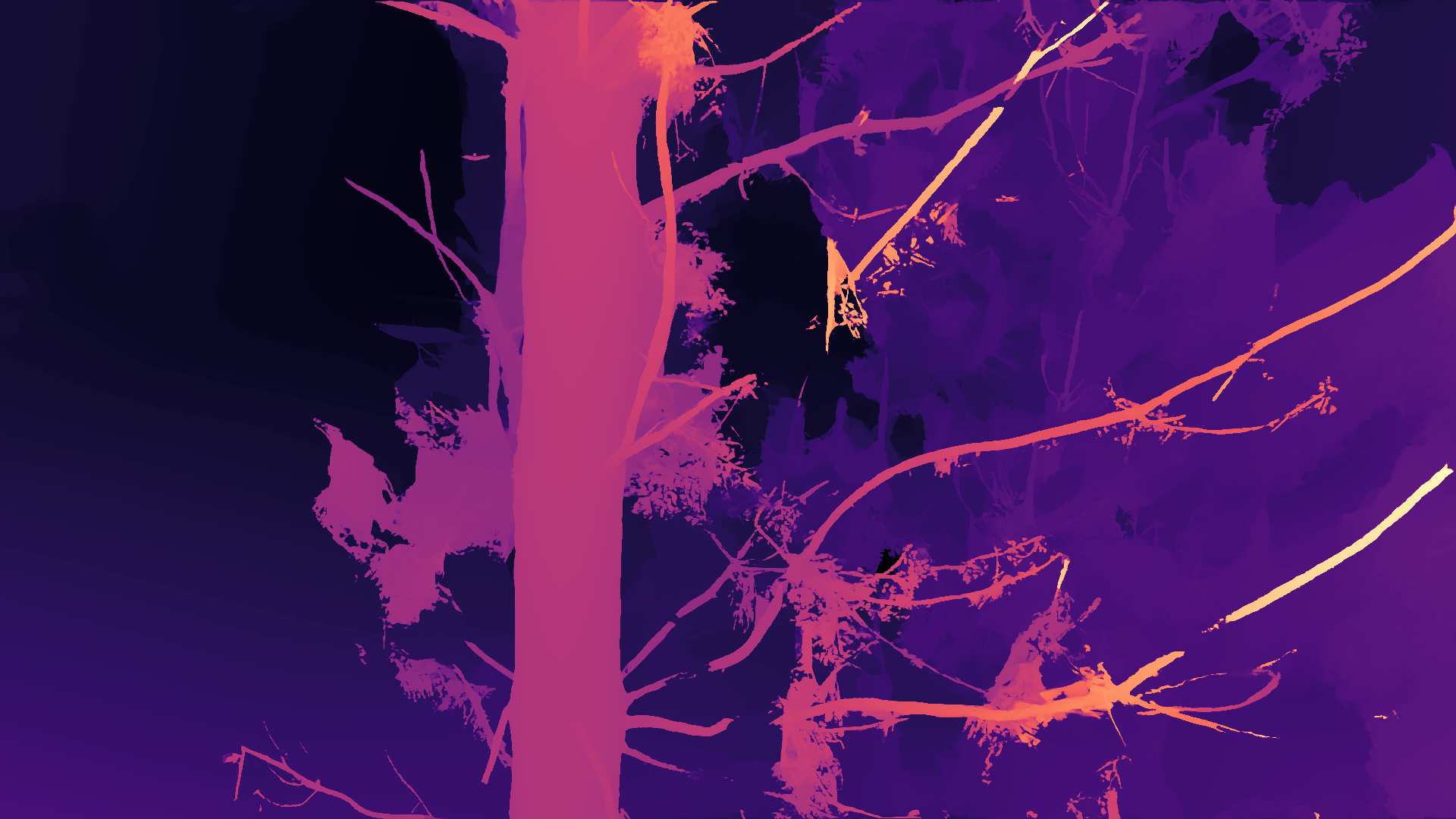}
            \caption{IGEV++ prediction}
        \end{subfigure}
        \caption{Qualitative comparison on Scene 5128 under variable lighting
        conditions. DEFOM maintains stable depth predictions. IGEV++ shows more
        texture detail but less consistent depth.}
        \label{fig:scene_5128}
    \end{figure*}

    \subsection{Qualitative Evaluation on Tree Branches Dataset}

    Based on the benchmark analysis, we apply DEFOM to our Tree Branches dataset
    to generate pseudo-ground-truth depth maps. For comparison, we also evaluate
    IGEV++, which shows the best D1 among iterative methods on Middlebury (7.82\%).
    Figs.~\ref{fig:scene_3305}--\ref{fig:scene_5128} present qualitative
    comparisons on representative vegetation scenes.

    \textbf{Scene 3305} (Fig.~\ref{fig:scene_3305}): Dense foliage with overlapping
    branches. DEFOM produces spatially coherent disparity maps with consistent
    sky-region predictions. IGEV++ preserves finer branch details but exhibits noise
    in homogeneous areas.

    \textbf{Scene 4939} (Fig.~\ref{fig:scene_4939}): Extreme depth
    discontinuities. DEFOM generates clean depth transitions at occlusion boundaries.
    IGEV++ shows sharper edges but exhibits occasional artifacts near thin
    structures.

    \textbf{Scene 5128} (Fig.~\ref{fig:scene_5128}): Dappled illumination through
    canopy. DEFOM maintains stable predictions across shadow boundaries. IGEV++ produces
    more detailed texture but less consistent depth.

    \subsection{Establishing DEFOM as Gold Standard}

    Based on quantitative and qualitative evaluation, \textbf{DEFOM is selected
    as the gold-standard method} for generating pseudo-ground-truth. Key factors:
    (1) top-2 EPE on 3 of 4 benchmarks without catastrophic failures; (2) smoother
    disparity maps with fewer artifacts on vegetation scenes; (3) stable
    performance across diverse domains.

    \vspace{-1em}
    \section{Discussion}

    \subsection{Cross-Domain Generalization Analysis}

    Foundation models demonstrate superior cross-domain consistency through
    large-scale pretraining. \textbf{DEFOM achieves best ranked consistency} (average
    rank 1.75 across 4 benchmarks) with lowest CV (0.58), ideal for establishing
    reference baselines. BridgeDepth achieves lowest absolute error on ETH3D (0.23~px)
    but higher variance across domains.

    \textbf{DEFOM vs.\ IGEV++ Trade-offs}: DEFOM excels in smoothness and
    consistency---essential for ground truth generation---achieving 31\% lower EPE
    on Middlebury. IGEV++ offers better outlier control (D1: 7.82\% vs.\ 8.28\%),
    valuable for safety-critical detection but less suitable as reference baseline.

    \textbf{Critical Finding}: Classical methods' catastrophic failures (PSMNet:
    54.42\% D1 on Middlebury) versus modern methods' stable performance demonstrates
    that architectural advances significantly improve cross-domain robustness.

    \subsection{Implications for Forestry Applications}

    DEFOM's selection as gold-standard baseline enables future quantitative
    evaluation without expensive LiDAR acquisition. Future work will validate against
    LiDAR measurements on selected scenes.

    \subsection{Limitations and Future Work}

    \textbf{Limitations}: (1) Tree Branches dataset evaluation is qualitative
    only; DEFOM's selection relies on cross-domain consistency and visual assessment.
    (2) Single-dataset training may not represent optimal generalization. (3) Statistical
    significance testing would strengthen comparative claims.

    \textbf{Future Directions}: (1) Establish the Tree Branches dataset as a public
    benchmark with DEFOM pseudo-ground-truth. (2) Computational profiling for
    embedded UAV deployment. (3) Uncertainty quantification for safety-critical applications.

    \section{Conclusion}

    This paper presented the first systematic zero-shot evaluation of deep
    stereo matching methods for UAV forestry applications. We evaluated eight methods
    spanning four architectural paradigms---iterative refinement, foundation
    models, diffusion-based, and 3D CNN---across four standard benchmarks and a
    novel 5,313-pair Tree Branches dataset.

    Our evaluation revealed distinct cross-domain generalization patterns. Foundation
    models leveraging monocular depth priors achieved superior consistency, with
    DEFOM ranking 1st--2nd across all benchmarks (average rank 1.75). Iterative
    methods demonstrated moderate but stable performance, while classical 3D CNN
    methods exhibited catastrophic failures on out-of-distribution data,
    underscoring the importance of modern architectures for zero-shot deployment.

    Based on these findings, we established DEFOM as the gold-standard baseline for
    generating pseudo-ground-truth on the Tree Branches dataset. This enables
    quantitative benchmarking of stereo methods on real-world vegetation scenes
    without expensive LiDAR annotation---addressing a critical need for
    developing autonomous UAV pruning systems that require centimeter-level
    depth accuracy.

    The Tree Branches dataset, along with DEFOM pseudo-ground-truth, will be
    publicly released to facilitate future research in UAV-based forestry
    automation.

    \section*{Acknowledgments}

    This research was supported by the Royal Society of New Zealand Marsden Fund
    and the Ministry of Business, Innovation and Employment. We thank the forestry
    research stations for data collection access and our annotators for ground truth
    generation.

    \bibliographystyle{IEEEtran}
    
\end{document}